\documentclass[letterpaper]{article} 
\usepackage{aaai2027}  
\usepackage[hyphens]{url}  
\usepackage{graphicx} 
\usepackage{natbib}  
\usepackage{caption} 
\usepackage{algorithm}
\usepackage{algorithmic}
\usepackage{graphicx}
\usepackage{amsmath}
\usepackage{booktabs}
\usepackage{multirow}
\usepackage[most]{tcolorbox}
\usepackage{enumitem}
\usepackage[table]{xcolor}
\usepackage[most]{tcolorbox}
\usepackage{listings}
\usepackage{xcolor}

\lstdefinestyle{promptstyle}{
    basicstyle=\fontfamily{ptm}\selectfont\scriptsize,
    columns=fullflexible,
    keepspaces=true,
    breaklines=true,
    breakindent=0pt,
    postbreak={},
    showstringspaces=false,
    numbers=none,
    xleftmargin=0pt,
    numbersep=0pt,
    tabsize=2,
    aboveskip=0pt,
    belowskip=0pt
}

\newtcblisting{promptbox}[1]{
    enhanced,
    listing only,
    width=\columnwidth,
    colback=gray!8,
    colframe=black!70,
    colbacktitle=black!75,
    coltitle=white,
    boxrule=0.8pt,
    arc=2pt,
    outer arc=2pt,
    left=2mm,
    right=2mm,
    top=1.5mm,
    bottom=1.5mm,
    toptitle=1mm,
    bottomtitle=1mm,
    lefttitle=2mm,
    fonttitle=\fontfamily{ptm}\selectfont\bfseries\normalsize,
    title={#1},
    listing options={style=promptstyle}
}
\newtcolorbox{prompttextbox}[1]{
    enhanced,
    width=\columnwidth,
    colback=gray!8,
    colframe=black!70,
    colbacktitle=black!75,
    coltitle=white,
    boxrule=0.8pt,
    arc=2pt,
    outer arc=2pt,
    left=2mm,
    right=2mm,
    top=1.5mm,
    bottom=1.5mm,
    toptitle=1mm,
    bottomtitle=1mm,
    lefttitle=2mm,
    fonttitle=\fontfamily{ptm}\selectfont\bfseries\normalsize,
    fontupper=\fontfamily{ptm}\selectfont\small,
    title={#1}
}

\newcommand{\ptag}[1]{\texttt{\textless #1\textgreater}}
\usepackage{newfloat}
 \usepackage{booktabs}
 \usepackage{subcaption}
 \usepackage{multirow}
 \usepackage{amssymb}   
 \usepackage{graphicx}  
\usepackage{listings}
\DeclareCaptionStyle{ruled}
{labelfont=normalfont,labelsep=colon,strut=off} 
\floatstyle{ruled}
\newfloat{listing}{tb}{lst}{}
\floatname{listing}{Listing}

\usepackage{booktabs}
\usepackage{amssymb}
\usepackage{arydshln}
\title{Mind the Divergence: Divergence-Guided Reliability-Aware Fusion and Adaptive LLM Prompting for Conversational Emotion Recognition}
\author{
    Jingwen Wang\textsuperscript{\rm 1,2},
    Jiao Li\textsuperscript{\rm 2},
    Yuhang Qian\textsuperscript{\rm 4},
    Yuyin Wang\textsuperscript{\rm 3},
    Min Zhang\textsuperscript{\rm 1}\corresponding
}

\affiliations{
    \textsuperscript{\rm 1}East China Normal University,
    \textsuperscript{\rm 2}Shanghai University, 
    \textsuperscript{\rm 3}Xidian University, \\
    \textsuperscript{\rm 4} Nanjing University of Aeronautics and Astronautics, \, 755792775@shu.edu.cn, mzhang@cs.ecnu.edu.cn
}
\begin{document}

\maketitle

\begin{abstract}
Multimodal emotion recognition in conversations (MERC) integrates textual, acoustic, and visual cues to identify utterance-level emotions. However, existing approaches generally do not explicitly adapt multimodal processing to the degree of inter-modal predictive disagreement. In real conversations, semantic incongruity, concealed emotion, noisy observations, and ambiguity can affect modality-specific emotional evidence differently, while the reliability of individual modalities varies across samples. Rather than identifying each source of inconsistency separately, we model their common consequence for emotion recognition as predictive disagreement among unimodal emotion distributions over the same emotion-label set. Such disagreement makes indiscriminate fusion prone to propagating unreliable or competing evidence. Inspired by the Brunswik Lens Model, we propose MiD-ERC, a divergence-centric framework that quantifies inter-modal predictive disagreement using a normalized divergence score and estimates sample-specific modality reliability. MiD-ERC uses these signals to coordinate divergence-guided cross-modal gating, divergence-modulated reliability weighting, and divergence-conditioned adaptive LLM prompting. The framework is trained end-to-end using only standard emotion annotations, without requiring additional conflict labels, auxiliary data construction, or external retrieval modules. Experiments on IEMOCAP and MELD demonstrate that MiD-ERC achieves state-of-the-art performance, with particularly pronounced improvements on high-divergence samples.
\end{abstract}

\section{Introduction}

Early research in ERC primarily focused on textual information
\cite{majumder2019dialoguernn,ghosal2019dialoguegcn,zhong2019knowledge,deng2021survey}.
More recent multimodal approaches incorporate acoustic and visual cues to complement textual semantics
\cite{chudasama2022m2fnet,li2022emocaps,lian2023mer}.
However, incorporating more modalities does not necessarily lead to more reliable emotion recognition.
In real conversations, semantic incongruity, concealed emotion, noisy observations, and ambiguity can affect different modalities in different ways.
For MERC, a source-agnostic abstraction of these heterogeneous inconsistencies is predictive disagreement among unimodal emotion distributions over the same emotion-label set.
Such disagreement makes indiscriminate fusion risky, because unreliable or competing modality-specific evidence may distort the resulting multimodal representation.
\par

\begin{figure}[!t]
    \centering
    \includegraphics[width=0.95\linewidth]{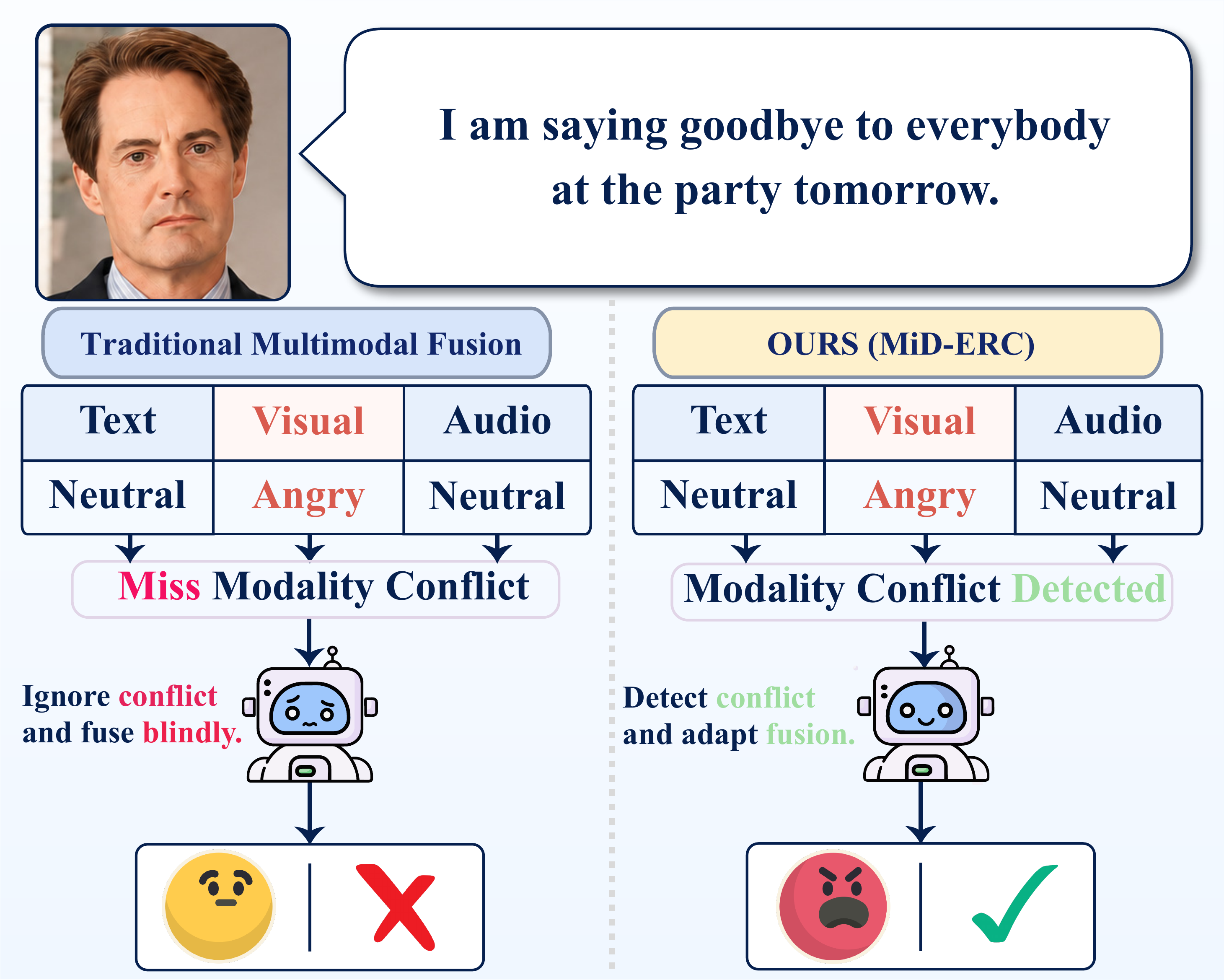}
    \caption{Orson \cite{akass2019reading} displays an angry facial expression while speaking in a relatively calm tone. MiD-ERC adapts multimodal fusion according to the resulting disagreement.}
    \label{fig:orson_drawio}
\end{figure}

In parallel, the emergence of LLMs has advanced ERC through strong contextual modeling and instruction-following capabilities
\cite{lei2023instructerc,xue2024bioserc,li2026llms}.
Nevertheless, existing multimodal and LLM-based ERC methods remain limited in several respects:
(1) \textbf{Lack of predictive-disagreement-aware multimodal fusion}.
Current multimodal systems do not explicitly coordinate their fusion behavior according to the degree of inter-modal predictive disagreement and sample-specific modality reliability.
Although Causal-ERC \cite{jing2026causal} integrates textual, acoustic, and visual features, its fusion behavior is not explicitly conditioned on predictive divergence.
Consequently, strongly disagreeing modality-specific evidence may still be integrated without sufficient control, allowing unreliable or ambiguous cues to distort the final multimodal representation.
(2) \textbf{Confinement to the textual modality}.
Many LLM-based approaches, including InstructERC \cite{lei2023instructerc}, BiosERC \cite{xue2024bioserc}, and PRC-Emo \cite{li2026llms}, primarily rely on text and therefore cannot fully exploit complementary acoustic and visual evidence.
(3) \textbf{Reliance on external components and auxiliary data construction}.
Existing LLM-based ERC methods often introduce additional engineering complexity beyond standard emotion annotations.
InstructERC uses SBERT-based retrieval for few-shot demonstrations
\cite{lei2023instructerc,reimers2019sentence},
BiosERC generates speaker biographies
\cite{xue2024bioserc},
and PRC-Emo introduces explicit and implicit interpretations together with a large retrieval repository
\cite{li2026llms}.
Although these components can provide useful contextual information, they increase system complexity and may introduce additional sources of error, whereas end-to-end learning from standard emotion annotations provides a simpler and more practical alternative.
\par
To address these challenges, we draw inspiration from the Brunswik Lens Model
\cite{brunswik2023perception,scherer2003vocal}.
The Lens Model views a speaker's internal emotional state as a distal variable that must be inferred from multiple proximal cues with different predictive validities.
Accurate recognition therefore depends on assessing the usefulness of individual cues and weighting them accordingly.
In multimodal conversations, textual, acoustic, and visual signals serve as such proximal cues, while their predictive reliability may vary across samples.
Translating this principle into a computational framework, we propose MiD-ERC, a divergence-centric framework for MERC.
MiD-ERC uses inter-modal predictive divergence as a global disagreement signal and modality-specific predictive reliability as a complementary local signal.
Together, these signals regulate three stages of multimodal processing: cross-modal interaction through divergence-guided cross-modal gating, evidence aggregation through divergence-modulated reliability weighting, and final prediction through divergence-conditioned adaptive LLM prompting.
When unimodal emotion predictions strongly disagree, weighted aggregation alone may be insufficient to determine how competing modality-specific evidence should be interpreted in context.
We therefore introduce a divergence-conditioned relation prompt with two evidence-use policies, guiding the LLM to adaptively use fused and modality-specific evidence for the final prediction.
\par

Our contributions are summarized as follows:
\begin{itemize}
    \item We introduce inter-modal predictive divergence as a continuous, source-agnostic control signal that requires no conflict-specific annotations.

    \item We develop divergence-guided cross-modal gating and divergence-modulated reliability weighting to regulate information exchange and evidence aggregation.

    \item We propose divergence-conditioned adaptive LLM prompting to adjust the use of fused and modality-specific evidence under different divergence levels.

    \item Experiments on IEMOCAP and MELD demonstrate state-of-the-art (SOTA) performance and show larger improvements on high-divergence samples.
\end{itemize}
\section{Related Work}
\subsection{Non-LLM MERC Methods}
Recent MERC methods improve multimodal representation learning and contextual modeling from different perspectives. CFN-ESA \cite{li2024cfn} adopts text-centered cross-modal interaction and emotion-shift awareness to exploit complementary acoustic and visual cues. LSDGNN+ICL \cite{li2025long} models long- and short-range conversational dependencies through parallel directed graphs and introduces curriculum learning for difficult and imbalanced samples. GraphSmile \cite{li2025tracing} alternates inter- and intra-modal graph propagation while modeling sentiment dynamics, whereas ConxGNN \cite{van2025effective} combines multi-scale heterogeneous graphs and hypergraphs to capture contextual and higher-order multimodal relations. GECP \cite{zhao2026graph} further learns category-level emotion consensus prototypes and aligns modality-specific representations through speaker-guided contrastive learning. Despite their effectiveness, these methods remain non-LLM architectures and therefore do not leverage the contextual interpretation and instruction-following capabilities of LLMs for final emotion prediction. In contrast, MiD-ERC integrates multimodal evidence into an LLM-based decision framework.\par
\subsection{LLM-Based MERC Methods}
In general emotion recognition, recent LLM-based approaches have begun to incorporate multimodal cues. AffectGPT \cite{lian2025affectgpt} and Emotion-LLaMAv2 \cite{peng2026emotion} encode acoustic and visual information and perform attention-based pre-fusion before LLM inference. However, the same pre-fusion strategy is applied across samples without explicitly adapting cross-modal interaction to the degree of modality disagreement. Moreover, these methods primarily focus on general clip-level emotion understanding rather than conversational ERC, and therefore do not explicitly model dialogue history, speaker dynamics, or cross-turn emotional dependencies, leading to suboptimal performance in MERC. Among LLM-based ERC methods, Causal-ERC is, to our knowledge, the only work integrating multimodal features into an LLM at the representation level. It combines speaker-aware contextual encoding, bidirectional cross-modal interaction, and a prompting mechanism based on the Peak-End Rule \cite{kahneman1993more,epstein1994integration}. Nevertheless, its fusion and prompting policies are not explicitly conditioned on predictive disagreement or sample-specific modality reliability. MiD-ERC addresses this limitation by adapting cross-modal interaction, evidence aggregation, and LLM prompting according to predictive divergence and modality reliability.
\begin{figure*}[htbp]
    \centering
    \includegraphics[width=\textwidth]{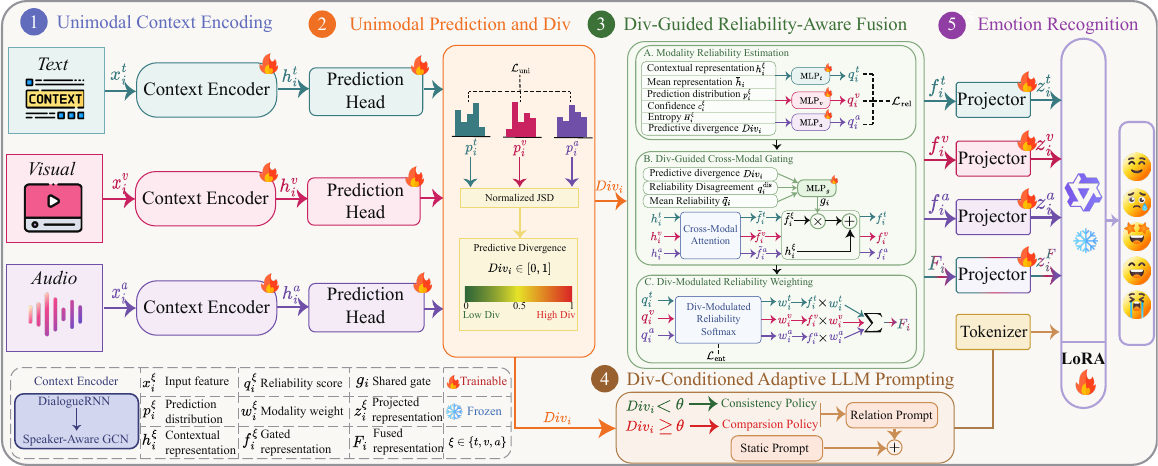} 
    \caption{Schematic diagram of the MiD-ERC framework.}
    \label{fig:mid_erc_double}
\end{figure*}
\section{MiD-ERC Framework}
We propose MiD-ERC, a divergence-centric framework for MERC, as illustrated in Figure~\ref{fig:mid_erc_double}. MiD-ERC models heterogeneous cross-modal inconsistencies through their common consequence for emotion recognition: predictive disagreement among unimodal emotion distributions. It quantifies this disagreement using a normalized inter-modal predictive divergence score and complements it with sample-specific modality reliability estimates. Together, these signals coordinate three stages of multimodal processing: divergence-guided cross-modal gating regulates information exchange, divergence-modulated reliability weighting controls evidence aggregation, and divergence-conditioned adaptive LLM prompting adjusts how the LLM uses fused and modality-specific evidence. Rather than identifying the specific source of disagreement, MiD-ERC adapts the overall processing pipeline when unimodal predictions provide competing emotional evidence. The following subsections describe these components in detail.
\subsection{Task Definition}

Given a dialogue consisting of $T$ utterances $\{u_1, u_2, \dots, u_T\}$ with corresponding speaker identities $\{s_1, s_2, \dots, s_T\}$, each utterance $u_i$ is associated with three modality features: text $x_i^t$, audio $x_i^a$, and visual $x_i^v$. The goal of MERC is to predict the emotion label $y_i \in \{1, 2, \dots, K\}$ for each utterance $u_i$, leveraging both the multimodal signals and the conversational context.
\subsection{Unimodal Context Encoding}

Following the Lens Model, each modality provides a proximal cue for inferring the speaker's internal emotional state. To obtain context-enriched unimodal representations, MiD-ERC encodes each modality independently through two stages.

\paragraph{Temporal context modeling.}
For each modality $\xi \in \{t,a,v\}$, a modality-specific DialogueRNN~\cite{majumder2019dialoguernn} captures temporal dependencies across the dialogue:
\begin{equation}
r_i^\xi =
\mathrm{DialogueRNN}_\xi
\left(x_i^\xi;\mathbf{sp}_i\right)
\end{equation}
where $\mathbf{sp}_i$ denotes the speaker embedding of utterance $i$, enabling the encoder to account for speaker-specific emotional dynamics.

\paragraph{Speaker-aware causal graph propagation.}
DialogueRNN summarizes the dialogue history through recurrent states, whereas DialogueGCN~\cite{ghosal2019dialoguegcn} permits bidirectional information propagation involving future utterances. To maintain causal processing, we introduce a single-layer Speaker-Aware Causal GCN that connects node $i$ only to preceding utterances within a historical window $W$. Separate transformations are applied to intra-speaker and inter-speaker edges to model speaker-specific emotional continuity and cross-speaker influence, respectively:
\begin{equation}
h_i^\xi =
\mathrm{LayerNorm}\!\left(
r_i^\xi +
\mathrm{Dropout}\!\left(
u_i^\xi + \bar{m}_i^\xi
\right)
\right)
\end{equation}
where $u_i^\xi$ denotes the self-projection of the target node, and $\bar{m}_i^\xi$ is the historical context aggregated over the neighborhood $\mathcal{N}_i$ using attention weights $\alpha_{ji}$. Detailed formulations and parameter definitions are provided in the Appendix A.1. The resulting representations $h_i^t$, $h_i^a$, and $h_i^v$ are used for unimodal prediction, divergence estimation, and subsequent adaptive multimodal processing.
\subsection{Unimodal Prediction and Divergence}
\label{sec:divergence}

\paragraph{Unimodal prediction.}
Each modality independently predicts an emotion distribution through a modality-specific classification head:
\begin{equation}
p_i^\xi =
\mathrm{softmax}\left(\mathrm{MLP}_\xi(h_i^\xi)\right)
\end{equation}
The three unimodal heads are supervised using an auxiliary cross-entropy loss:
\begin{equation}
\mathcal{L}_{\mathrm{uni}}
=
\frac{1}{3}
\sum_{\xi \in \{t,a,v\}}
\mathrm{CE}(p_i^\xi,y_i)
\end{equation}
This auxiliary supervision aligns unimodal predictions within a shared emotion-label space, enabling their discrepancies to serve as a source-agnostic signal for adapting subsequent multimodal processing.

\paragraph{Predictive divergence.}
\paragraph{Predictive divergence.}
To quantify inter-modal predictive disagreement, we compute a normalized Jensen-Shannon divergence (JSD)~\cite{lin1991divergence}:
\begin{equation}
\mathrm{Div}_i
=
\frac{1}{3\log 3}
\sum_{\xi \in \{t,a,v\}}
\mathrm{KL}
\left(
p_i^\xi \parallel \bar{p}_i
\right)
\end{equation}
where $\bar{p}_i$ denotes the mean emotion distribution across the three modalities, and $\log 3$ corresponds to the tight upper bound of the generalized
JSD among three equally weighted modality distributions. Dividing by
$\log 3$ therefore normalizes the divergence to $[0,1]$, independently
of the number of emotion classes $K$. The resulting predictive divergence serves as a global disagreement signal for modality reliability estimation, divergence-guided cross-modal gating, divergence-modulated reliability weighting, and divergence-conditioned adaptive LLM prompting.

\subsection{Modality Reliability Estimation}
While the divergence score $\mathrm{Div}_i$ quantifies the overall degree of inter-modal disagreement, it does not indicate which modality is more reliable for a given sample. In the Brunswik Lens Model, this distinction is characterized by the ecological validity of individual cues. Inspired by this concept, we estimate a sample-specific reliability score for each modality. Specifically, a modality-specific $\mathrm{MLP}_\xi$ followed by a sigmoid activation computes the reliability score $q_i^\xi$:

\begin{equation}
q_i^\xi = \sigma\left(\mathrm{MLP}_\xi\left(\left[h_i^\xi; \bar{h}_i; p_i^\xi; c_i^\xi; H_i^\xi; {\mathrm{Div}}_i\right]\right)\right)
\end{equation}

where $h_i^\xi$, $\bar{h}_i$, and $p_i^\xi$ denote the contextual representation, the mean representation across the three modalities, and the unimodal emotion prediction distribution, respectively. The scalars $c_i^\xi$, $H_i^\xi$, and $\mathrm{Div}_i$ represent the maximum prediction confidence, prediction entropy, and inter-modal predictive divergence. Prior studies \cite{li2026geometric,lin2026v2vloc} have demonstrated the effectiveness of concatenating task-relevant scalar information with high-dimensional representations, and our experiments further validate this design in Section~4.6. For each sample, we define the true reliability of modality $\xi$ as $p_i^\xi(y_i)$, namely, the probability assigned by its unimodal classifier to the ground-truth label $y_i$, and use it to supervise the reliability estimator through the following mean squared error objective:
\begin{equation}
\mathcal{L}_{\text{rel}}
=
\frac{1}{3}
\sum_{\xi \in \{t,a,v\}}
\mathrm{MSE}
\left(
q_i^\xi,
\mathrm{sg}\left(p_i^\xi(y_i)\right)
\right)
\end{equation}
where $\mathrm{sg}(\cdot)$ denotes the stop-gradient operation, which prevents the reliability loss from affecting the unimodal classifiers.
\subsection{Div-Guided Cross-Modal Gating}
To model sample-specific interactions across heterogeneous modalities, we adopt a parallel cross-modal attention mechanism \cite{tsai2019multimodal}. Taking the textual modality as an example, the cross-modally enhanced feature is computed as:
\begin{equation}
\tilde f_i^t= \mathrm{MLP}_c \left[ \mathrm{CA}(h_i^t,\mathbf H_i^a); \mathrm{CA}(h_i^t,\mathbf H_i^v) \right] 
\end{equation}
where $\mathrm{CA}(\cdot)$ denotes cross-attention, $\mathbf{H}_i^a$ and $\mathbf{H}_i^v$ denote the acoustic and visual hidden-state sequences, respectively, and $\mathrm{MLP}_{c}$ integrates the concatenated attention outputs. The enhanced features $\tilde{f}_i^a$ and $\tilde{f}_i^v$ are obtained analogously.

Uniform cross-modal interaction may introduce inconsistent or less reliable evidence. We therefore regulate information exchange using a divergence-guided gate based on inter-modal predictive divergence and modality reliability. Let $\bar{q}_i$ denote the average reliability across the three modalities. The reliability disagreement is defined as:

\begin{equation}
q_i^{\mathrm{dis}}
=
\frac{1}{3}
\sum\limits_{\xi\in\{t,a,v\}}
\left|
q_i^\xi-\bar{q}_i
\right|
\end{equation}

The gate is then computed as:

\begin{equation}
g_i =
\sigma
\left(
\mathrm{MLP}_{g}
\left(
\left[
\mathrm{Div}_i;
\bar{q}_i;
q_i^{\mathrm{dis}}
\right]
\right)
\right)
\end{equation}

Here, $\mathrm{Div}_i$ measures the overall predictive disagreement, $\bar{q}_i$ reflects the overall reliability of the multimodal evidence, and $q_i^{\mathrm{dis}}$ captures the reliability variation across modalities. A higher $\bar{q}_i$ promotes cross-modal interaction, while a higher $q_i^{\mathrm{dis}}$ reflects greater reliability imbalance and enables adaptive control of cross-modal information injection. These effects are analyzed in Section 4.6.

Finally, the enhanced feature is incorporated through a gated residual connection:
\begin{equation}
f_i^\xi =
h_i^\xi +
g_i \cdot \tilde{f}_i^\xi
\end{equation}
\subsection{Div-Modulated Reliability Weighting}
We use a divergence-dependent temperature to control the concentration of modality weights:

\begin{equation}
\tau_i=\tau_0\left(1+\alpha\,\mathrm{Div}_i\right)
\end{equation}

where $\tau_0$ is the base temperature and $\alpha$ controls its sensitivity to predictive divergence. The modality weights are computed from the reliability scores as:

\begin{equation}
w_i^\xi=
\frac{\exp\left(q_i^\xi/\tau_i\right)}
{\sum\limits_{\xi'\in\{t,a,v\}}\exp\left(q_i^{\xi'}/\tau_i\right)}
\end{equation}

$q_i^\xi$ determines the relative importance of each modality, while $\mathrm{Div}_i$ controls the concentration of the weight distribution through $\tau_i$. A higher $\mathrm{Div}_i$ produces smoother weights, preserving competing evidence for the LLM, consistent with the Bayesian causal inference perspective discussed in the Appendix A.2 \cite{mazzi2025prior}. The weighted multimodal representation is then computed as:

\begin{equation}
F_i=
\mathrm{LayerNorm}
\left(
\sum_{\xi\in\{t,a,v\}}w_i^\xi f_i^\xi
\right)
\end{equation}

We further introduce a divergence-weighted entropy regularization term to discourage single-modality dominance in high-divergence samples:

\begin{equation}
\mathcal{L}_{\mathrm{ent}}
=
\mathrm{Div}_i
\sum_{\xi\in\{t,a,v\}}
w_i^\xi\log\left(w_i^\xi+\epsilon\right)
\end{equation}
\subsection{Div-Conditioned Prompting}
Before prompting, we project the reliability-weighted representation $F_i$ and gated modality-specific representations $f_i^\xi$ into the LLM embedding space:

\begin{equation}
z_i^F = W_F F_i + b_F
\end{equation}
\begin{equation}
z_i^\xi = W_\xi f_i^\xi + b_\xi
\end{equation}
Based on $\mathrm{Div}_i$ and threshold $\theta$, low-divergence samples adopt a Consistency Policy, treating $F_i$ as primary evidence and $f_i^\xi$ as supporting evidence. High-divergence samples adopt a Comparison Policy, prompting the LLM to contrast modality-specific evidence with the fused representation and dialogue context. The selected Relation Prompt is combined with the Static Prompt for a single LLM forward pass. Detailed templates are provided in the Appendix B.
\subsection{Training Objective}
The main classification loss is given by:
\begin{equation}\mathcal{L}_{\text{LLM}} = \mathrm{CE}(\hat{y}_i, y_i)
\end{equation}
where $\hat{y}_i$ represents the predicted probability distribution generated by the LLM, and $y_i$ signifies the ground-truth label. Consequently, the total loss function is given by:
\begin{equation}
\mathcal{L} = \mathcal{L}_{\text{LLM}} + \lambda_{\text{uni}} \mathcal{L}_{\text{uni}} + \lambda_{\text{rel}} \mathcal{L}_{\text{rel}} + \lambda_{\text{ent}} \mathcal{L}_{\text{ent}}
\end{equation}
where $\lambda_{\text{uni}}$, $\lambda_{\text{rel}}$, and $\lambda_{\text{ent}}$ denote the balancing hyperparameters.
\section{Experiments}
\subsection{Datasets}
We evaluate MiD-ERC on the IEMOCAP \cite{busso2008iemocap} and MELD \cite{poria2019meld} datasets, with statistics summarized in Table 1. As the original multimodal formats are incompatible with standard LLM inputs, we convert the data into an instruction-following format for training and inference. Further details on the datasets and the conversion process are provided in the Appendix C.
\begin{table}[h]
    \centering
   
    \label{tab:dataset_stats} 
    \resizebox{\columnwidth}{!}{
    \begin{tabular}{c ccc ccc c}
        \toprule
        \multirow{2}{*}{Dataset} & \multicolumn{3}{c}{Conversations} & \multicolumn{3}{c}{Utterances} & \multirow{2}{*}{Classes} \\
        \cmidrule(lr){2-4} \cmidrule(lr){5-7}
        & train & val & test & train & val & test & \\
        \midrule
        IEMOCAP & \multicolumn{2}{c}{120} & 31 & \multicolumn{2}{c}{5810} & 1623 & 6 \\
        MELD & 1039 & 114 & 280 & 9989 & 1109 & 2610 & 7 \\
        \bottomrule
    \end{tabular}
    }
     \caption{Statistics of the two datasets.}
\end{table}
\subsection{Baselines}
We compare our proposed method with several SOTA baselines, including DialogueGCN \cite{ghosal2019dialoguegcn}, CFN-ESA \cite{li2024cfn}, LSDGNN+ICL \cite{li2025long}, GraphSmile \cite{li2025tracing}, ConxGNN \cite{van2025effective}, GECP \cite{zhao2026graph}, InstructERC \cite{lei2023instructerc}, BiosERC \cite{xue2024bioserc}, PRC-Emo \cite{li2026llms}, and Causal-ERC \cite{jing2026causal}. We employ Accuracy (Acc) and Weighted-F1 (W-F1) as our evaluation metrics.
\subsection{Implementation Details}
All experiments are conducted on a single NVIDIA RTX 5880 Ada GPU with CUDA 12.1. We adopt Qwen2.5-7B-Instruct as the base LLM, and the experimental results on additional LLM backbones are provided in the Appendix E.1. To balance performance and parameter efficiency, we freeze the Qwen2.5-7B-Instruct backbone and apply Low-Rank Adaptation (LoRA) \cite{hu2022lora} to all its linear layers. Comprehensive training details and the complete set of empirically searched hyperparameters are provided in the Appendix D. All reported results are averaged over six random seeds.

\begin{table*}[t]
\centering
\setlength{\tabcolsep}{3pt}
\renewcommand{\arraystretch}{1.15}
\resizebox{\textwidth}{!}{%
\begin{tabular}{lccccccccccccccccc} 
\toprule
\multirow{2}{*}[-1ex]{Methods} & \multicolumn{8}{c}{IEMOCAP} & \multicolumn{9}{c}{MELD} \\
\cmidrule(lr){2-9} \cmidrule(lr){10-18}
 & Happy & Sad & Neutral & Angry & Excited & Frustrated & Acc & W-F1 & Neutral & Surprise & Fear & Sadness & Joy & Disgust & Anger & Acc & W-F1 \\
\midrule
$\bigstar$DialogueGCN$^{\flat}$ \cite{ghosal2019dialoguegcn} & 51.57 & 80.48 & 57.69 & 53.95 & 72.81 & 57.33 & 63.22 & 62.89 & 75.97 & 46.05 & - & 19.60 & 51.20 & - & 40.83 & 58.62 & 56.36 \\
$\bigstar$CFN-ESA$^{\sharp}$ \cite{li2024cfn} & 53.67 & 80.60 & 71.65 & 70.32 & 74.82 & 68.06 & 70.78 & 71.04 & 80.05 & 58.78 & 21.62 & 41.82 & 66.50 & 26.92 & 54.18 & 67.85 & 66.70 \\
$\bigstar$LSDGNN+ICL$^{\sharp}$ \cite{li2025long}    & 47.20 & 81.53 & 71.00 & 69.74 & 73.19 & 68.58 & 70.35 & 70.24 & 77.58 & 57.90 & 18.18 & 37.94 & 61.70 & 29.70 & 51.88 & 64.67 & 64.07  \\
$\bigstar$GraphSmile$^{\ddagger}$ \cite{li2025tracing}       & 60.14 & 81.38 & 68.48 & \textbf{70.45} & 77.60 & 66.02 & 70.98 & 71.00 & 62.99 & 51.96 & 32.14 & 28.85 & 39.24 & \textbf{56.62} & \textbf{79.95} & 66.67 & 65.46 \\
$\bigstar$ConxGNN$^{\sharp}$ \cite{van2025effective}       & - & - & - & - & - & - & 68.52 & 68.64 & - & - & - & - & - & - & - & 66.28 & 65.69 \\
$\bigstar$GECP$^{\ddagger}$ \cite{zhao2026graph}    & 42.47 & 79.24 & 65.82 & 69.03 & 75.84 & 61.93 & 67.16 & 67.04 & 78.36 & 58.67 & 21.05 & 41.64 & 64.09 & 26.92 & 53.00 & 67.01 & 65.33 \\
\midrule
$\clubsuit$$\bigstar$InstructERC$^{\sharp}$ \cite{lei2023instructerc}            & - & - & - & - & - & - & - & 71.39 & - & - & - & - & - & - & - & - & 69.15 \\
$\clubsuit$$\bigstar$BiosERC$^{\sharp}$ \cite{xue2024bioserc}        & - & - & - & - & - & - & - & 71.19 & - & - & - & - & - & - & - & - & 69.83 \\
$\clubsuit$$\bigstar$PRC-Emo$^{\sharp}$ \cite{li2026llms}              & - & - & - & - & - & - & 71.03 & 71.95 & - & - & - & - & - & - & - & - & 69.73 \\
Causal-ERC$^{\ddagger}$ \cite{jing2026causal}    & 58.90 & 82.96 & 72.86 & 67.97 & 70.41 & 69.79 & 71.60 & 71.46 & 82.06 & 62.43 & 25.81 & 42.32 & 65.64 & 37.93 & 59.32 & 70.08 & 69.02 \\
\midrule
$\bigstar$Qwen2.5-7B-Instruct$^{\ddagger}$   & \textbf{61.59} & 80.35 & 72.12 & 65.14 & 69.94 & 69.82 & 70.61 & 70.75 & 80.52 & 59.56 & 20.00 & 41.28 & 66.27 & 33.64 & 53.44 & 68.85 & 66.98 \\
MiD-ERC (Ours)
& 57.25 & \textbf{83.86} & \textbf{73.63} & 69.86
& \textbf{77.74} & \textbf{70.63}
& \textbf{73.44}{\scriptsize $\uparrow$2.83}
& \textbf{73.38}{\scriptsize $\uparrow$2.63}
& \textbf{82.83} & \textbf{62.73} & \textbf{36.36}
& \textbf{49.22} & \textbf{68.38} & 37.84 & 59.38
& \textbf{71.79}{\scriptsize $\uparrow$2.94}
& \textbf{70.60}{\scriptsize $\uparrow$3.62} \\
\bottomrule
\end{tabular}%
}
\caption{We compare the overall performance of various methods using Acc, W-F1, and per-class F1 scores. $^\sharp$, $^\flat$, and $^\ddagger$ denote results from original papers, MMDFN \cite{hu2022mm}, and our reproduction, respectively. $\clubsuit$ signifies external components or re-annotated datasets, while $\bigstar$ indicates open-source code availability.}
\label{tab:main_results}
\end{table*}
\subsection{Results and Discussion}
Table~\ref{tab:main_results} shows that MiD-ERC achieves the best overall performance, with 73.44\% Acc and 73.38\% W-F1 on IEMOCAP, and 71.79\% Acc and 70.60\% W-F1 on MELD. Compared with Qwen2.5-7B-Instruct using only task-specific SFT, it improves Acc by 2.83\% and 2.94\%, and W-F1 by 2.63\% and 3.62\% on the two datasets, respectively. It also exceeds Causal-ERC by 1.84\% and 1.71\% in Acc, and by 1.92\% and 1.58\% in W-F1, while outperforming the strongest non-LLM baseline, CFN-ESA, by 2.34\% and 3.90\% in W-F1. Moreover, MiD-ERC achieves the best F1 on nine of thirteen emotion classes, including gains of 10.55\% and 6.90\% over Causal-ERC on MELD Fear and Sadness.\par
\begin{table}[!t]
  \centering
  
  \label{tab:inference_time}
  \begin{tabular}{ccccc}
    \toprule
    \multirow{2}{*}{Methods} & \multicolumn{2}{c}{IEMOCAP} & \multicolumn{2}{c}{MELD} \\
    \cmidrule(lr){2-3} \cmidrule(lr){4-5}
    & Single & Total & Single & Total \\
    \midrule
    \textbf{MiD-ERC} & \textbf{0.18} & \textbf{292.14} & \textbf{0.15} & \textbf{391.50} \\
    Causal-ERC       & 0.34          & 551.82          & 0.28          & 730.80          \\
    \bottomrule
  \end{tabular}
  \caption{Inference time comparison.}
\end{table}
\begin{table}[!t]
  \centering
  \resizebox{\columnwidth}{!}{%
  \begin{tabular}{cccccccc} 
    \toprule
    \multirow{2}{*}[-1ex]{Methods} & \multicolumn{2}{c}{IEMOCAP} & \multicolumn{2}{c}{MELD} \\
    \cmidrule(lr){2-3} \cmidrule(lr){4-5}
    & Acc & W-F1 & Acc & W-F1 \\
    \midrule
    Ours                                 & 73.44 & 73.38 & 71.79 & 70.60 \\
    w/ T only                           & 71.47 & 71.35 & 70.46 & 69.15 \\
    w/ TA only                          & 72.58 & 72.51 & 70.88 & 69.55 \\
    w/o Speaker-Aware GCN            & 72.21 & 72.34 & 70.91 & 69.58 \\
    w/o $\mathcal{L}_{\text{uni}}$            & 71.04 & 71.05 & 71.11 & 69.72 \\
    w/o $\mathcal{L}_{\text{rel}}$        & 72.46 & 72.43 & 70.54 & 69.35 \\
    w/o $\mathcal{L}_{\text{ent}}$ & 71.72 & 71.85 & 70.61 & 69.52 \\
      w/o Relation Prompt   & 71.33 & 71.36 & 70.67 & 69.54 \\
         w/o $g_i$   & 72.02 & 72.12 & 70.65 & 69.16 \\
    w/o $F_i$   & 71.47 & 71.66 & 70.38 & 69.42 \\
   
    \bottomrule
  \end{tabular}%
  }
     \caption{Ablation study on IEMOCAP and MELD.}
\end{table}
Table 3 shows that MiD-ERC nearly halves inference time while achieving higher Acc and W-F1 on both datasets. This efficiency stems from lightweight unimodal divergence estimation and a single LLM forward pass, whereas Causal-ERC requires two passes per utterance.
\subsection{Ablation Study}
Table 4 reports component ablation studies validating their effectiveness. Section 4.6 will further detail the mechanisms of predictive divergence and reliability.

\textbf{Impact of multimodal information.} Text-only input drops W-F1 by 2.03\% and 1.45\% on IEMOCAP and MELD, respectively. Incorporating the acoustic modality recovers much of this gap, with the tri-modal model achieving the best overall performance. These results demonstrate the complementary contributions of acoustic and visual cues, supporting the use of multimodal evidence for MERC.

\textbf{Impact of speaker-aware GCN.} Removing the speaker-aware GCN drops W-F1 by 1.04\% on IEMOCAP and 1.02\% on MELD. Given the difficulty of the dataset, the decline on MELD is more significant. This demonstrates that explicitly distinguishing between intra-speaker and inter-speaker dependencies is critical as the number of interlocutors increases.

\textbf{Impact of auxiliary losses.} Removing $\mathcal{L}_{uni}$ is most damaging on IEMOCAP, dropping W-F1 by 2.33\%, alongside a 0.88\% decline on MELD. Since predictive divergence is computed from unimodal predictions, poor unimodal predictions undermine the reliability of subsequent divergence modeling. Removing $\mathcal{L}_{ent}$ and $\mathcal{L}_{rel}$ degrades performance across datasets, dropping W-F1 by up to 1.25\% on MELD. This confirms entropy regularization prevents single-modality dominance under high predictive divergence, while reliability supervision improves the estimation of predictive reliability for subsequent modality weighting.

\textbf{Impact of the Relation Prompt.} Replacing the Relation Prompt with the static prompt alone lowers W-F1 by 2.02\% on IEMOCAP and 1.06\% on MELD, demonstrating the benefit of adapting the LLM prompting strategy to the level of predictive divergence.\par
\textbf{Impact of the gate.} Removing gate $g_i$ drops W-F1 by 1.26\% on IEMOCAP and 1.44\% on MELD. Without it, cross-modal enhancement $\tilde{f}_i^\xi$ is added indiscriminately rather than modulated by inter-modal predictive divergence, preventing the model from adaptively regulating cross-modal information injection. The consistent degradation across both datasets confirms the effectiveness of divergence-guided cross-modal gating for cross-modal feature enhancement.
\par
\textbf{Impact of the fused representation.} Discarding the fused multimodal representation $F_i$ drops W-F1 by 1.72\% on IEMOCAP and 1.18\% on MELD. This consistent degradation shows that $F_i$ provides complementary global multimodal evidence beyond the modality-specific representations, benefiting the final emotion prediction.

\subsection{Divergence and Reliability Analysis}
This section analyzes how predictive divergence and reliability affect cross-modal interaction, modality weighting, and model performance, providing further insights into the internal mechanisms of MiD-ERC.\par
\textbf{Performance across divergence levels.} Figure 3 compares MiD-ERC and Causal-ERC across three $\text{Div}_i$ groups. MiD-ERC's advantage scales with divergence, with the W-F1 gap expanding to 2.87\% on IEMOCAP and 3.88\% on MELD for high-divergence samples, but diminishing or slightly regressing on low-divergence ones. These results support predictive divergence as an effective signal for adapting to cross-modal disagreement.\par
\begin{figure}[h]
    \centering
    \begin{subfigure}[b]{0.49\linewidth}
        \centering
        \includegraphics[width=\linewidth]{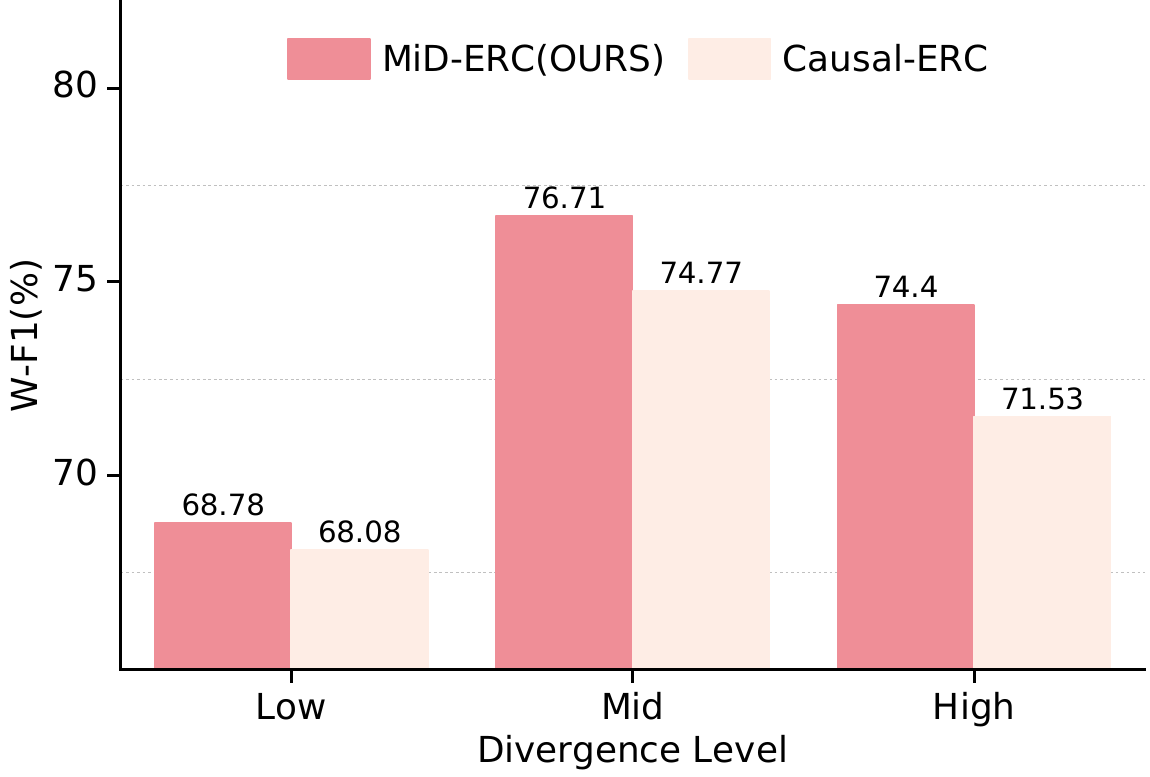}
        \caption{IEMOCAP} 
        \label{fig:sub_iemocap}
    \end{subfigure}
    \hfill
    \begin{subfigure}[b]{0.49\linewidth}
        \centering
        \includegraphics[width=\linewidth]{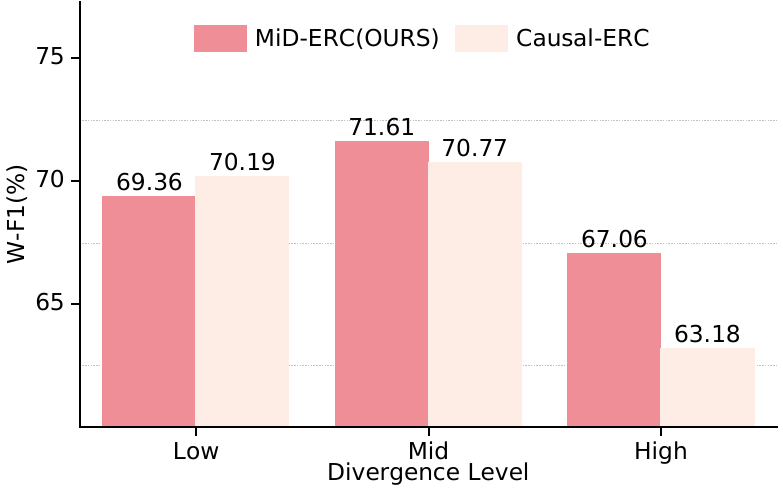}
        \caption{MELD} 
        \label{fig:sub_meld}
    \end{subfigure}
    
    \caption{Performance across divergence levels on IEMOCAP and MELD.}
    \label{fig:divergence}
\end{figure}
\begin{figure}[h]
    \centering
    \begin{subfigure}[b]{0.49\linewidth}
        \centering
        \includegraphics[width=\linewidth]{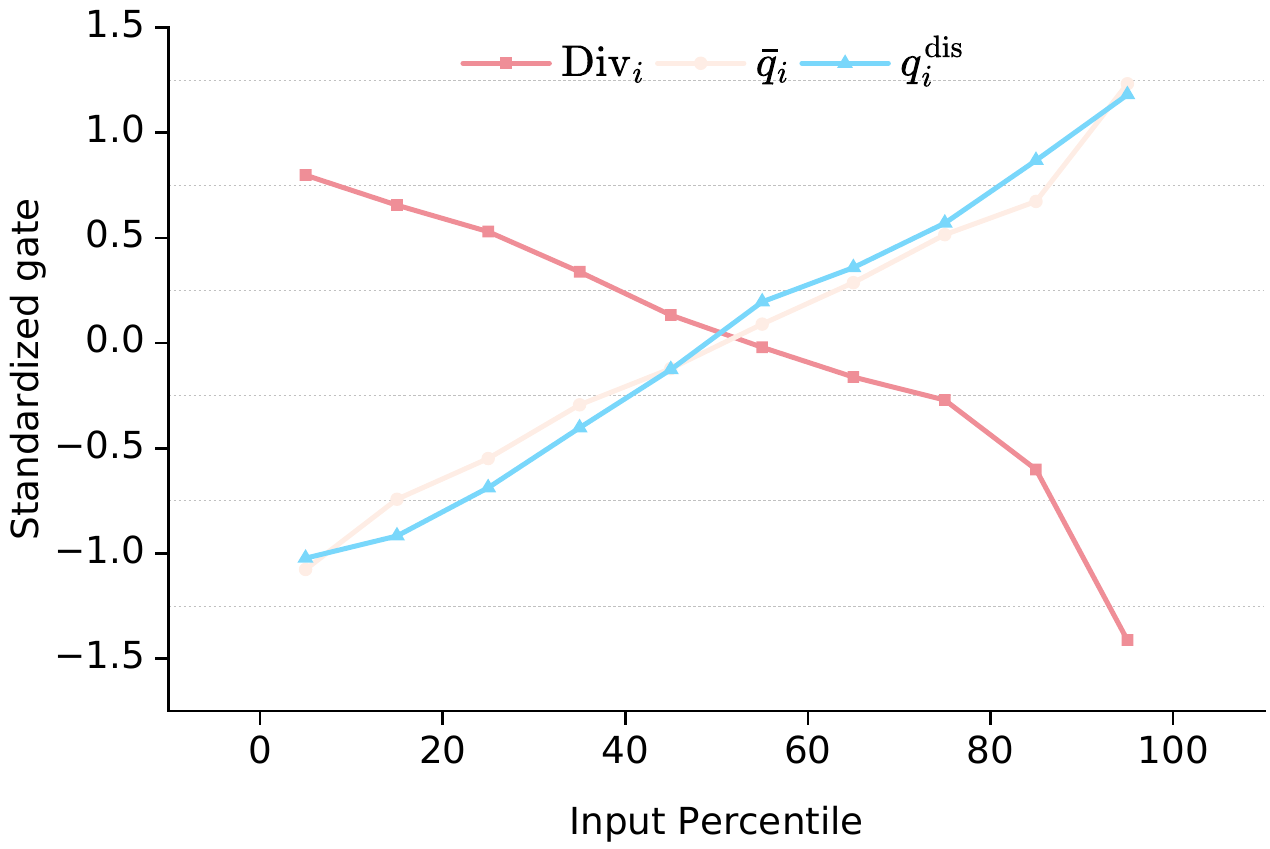}
        \caption{Input-binned gate output.} 
        \label{fig:sub_iemocap}
    \end{subfigure}
    \hfill
    \begin{subfigure}[b]{0.49\linewidth}
        \centering
        \includegraphics[width=\linewidth]{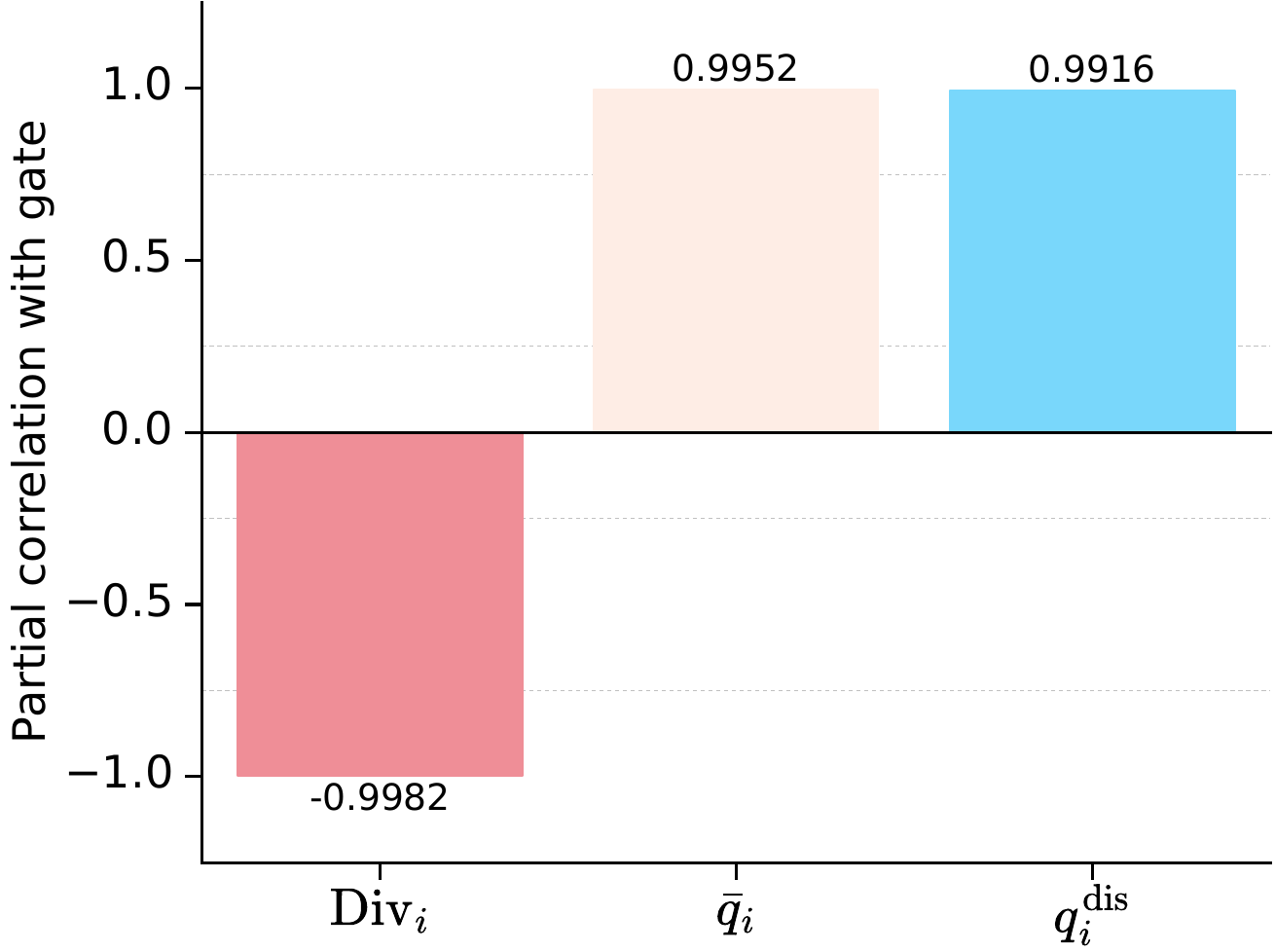}
        \caption{Input-$g_i$ partial correlation.} 
        \label{fig:sub_meld}
    \end{subfigure}
    
\caption{Learned gate $g_i$ versus its three inputs.}
    \label{fig:divergence}
\end{figure}
\begin{figure}[h]
    \centering
    \begin{subfigure}[b]{0.49\linewidth}
        \centering
        \includegraphics[width=\linewidth]{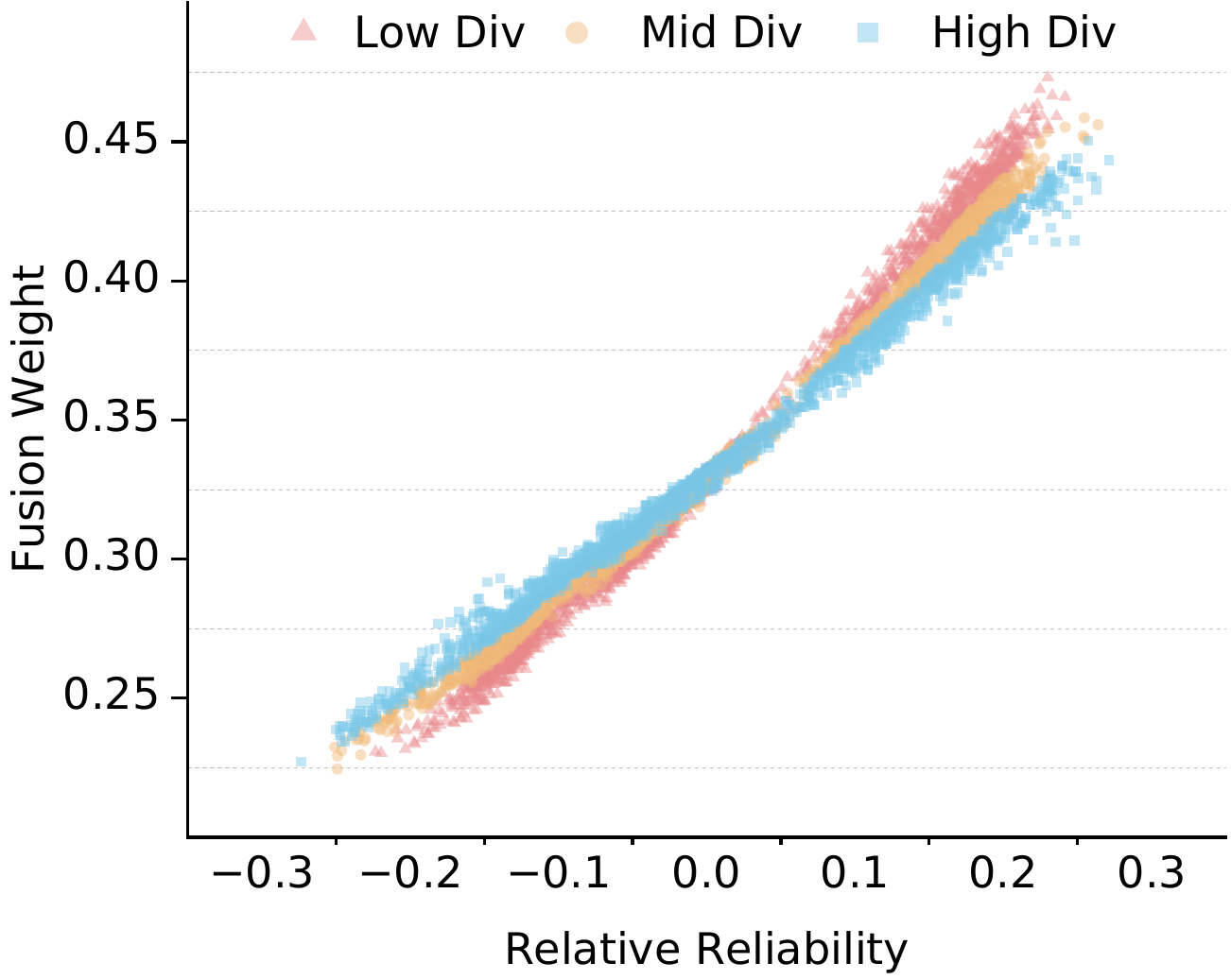}
        \caption{w/ $\mathcal{L}_{\text{ent}}$} 
        \label{fig:sub_iemocap}
    \end{subfigure}
    \hfill
    \begin{subfigure}[b]{0.49\linewidth}
        \centering
        \includegraphics[width=\linewidth]{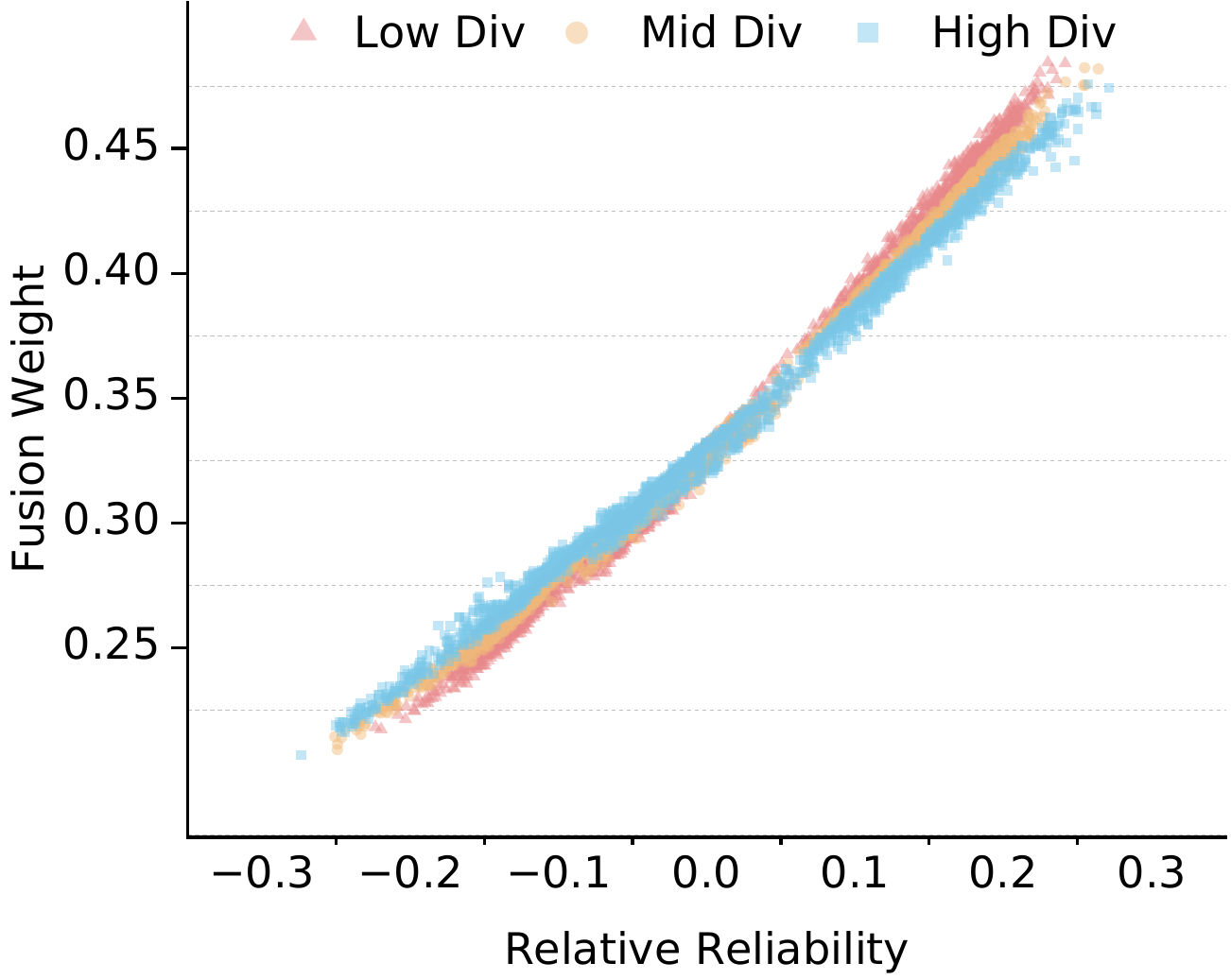}
        \caption{w/o $\mathcal{L}_{\text{ent}}$} 
        \label{fig:sub_meld}
    \end{subfigure}
    
\caption{Fusion weight $w_i^\xi$ versus relative reliability.}
    \label{fig:divergence}
\end{figure}
\textbf{Analysis of divergence-guided gate inputs.} We analyze the learned gate $g_i$ after the ablation study. Figure 4a shows that high predictive divergence suppresses the gate, while high $\bar{q}_i$ promotes information exchange and high $q_i^{\text{dis}}$ opens the gate under reliability imbalance. The partial correlations \cite{yule1907theory} in Figure 4b further shows that all three inputs contribute distinct information. These results demonstrate that $g_i$ interpretably regulates cross-modal enhancement using divergence and reliability cues.\par
\textbf{Analysis of fusion weight allocation.} Figure 5 shows that the fusion weight $w_i^\xi$ increases monotonically with relative reliability, indicating that our fusion mechanism assigns larger weights to more reliable modalities. As predictive divergence increases, the curves become flatter, which is consistent with the proposed temperature scaling. Moreover, removing $\mathcal{L}_{\text{ent}}$ produces sharper peaks, with most of the weight assigned to a single modality, while using it smooths the weights toward uniformity, especially high predictive divergence. This supports the role of entropy regularization in preventing over-dominance of a single modality.\par
\begin{figure}[!t]
    \centering
    \begin{subfigure}[b]{0.49\linewidth}
        \centering
        \includegraphics[width=\linewidth]{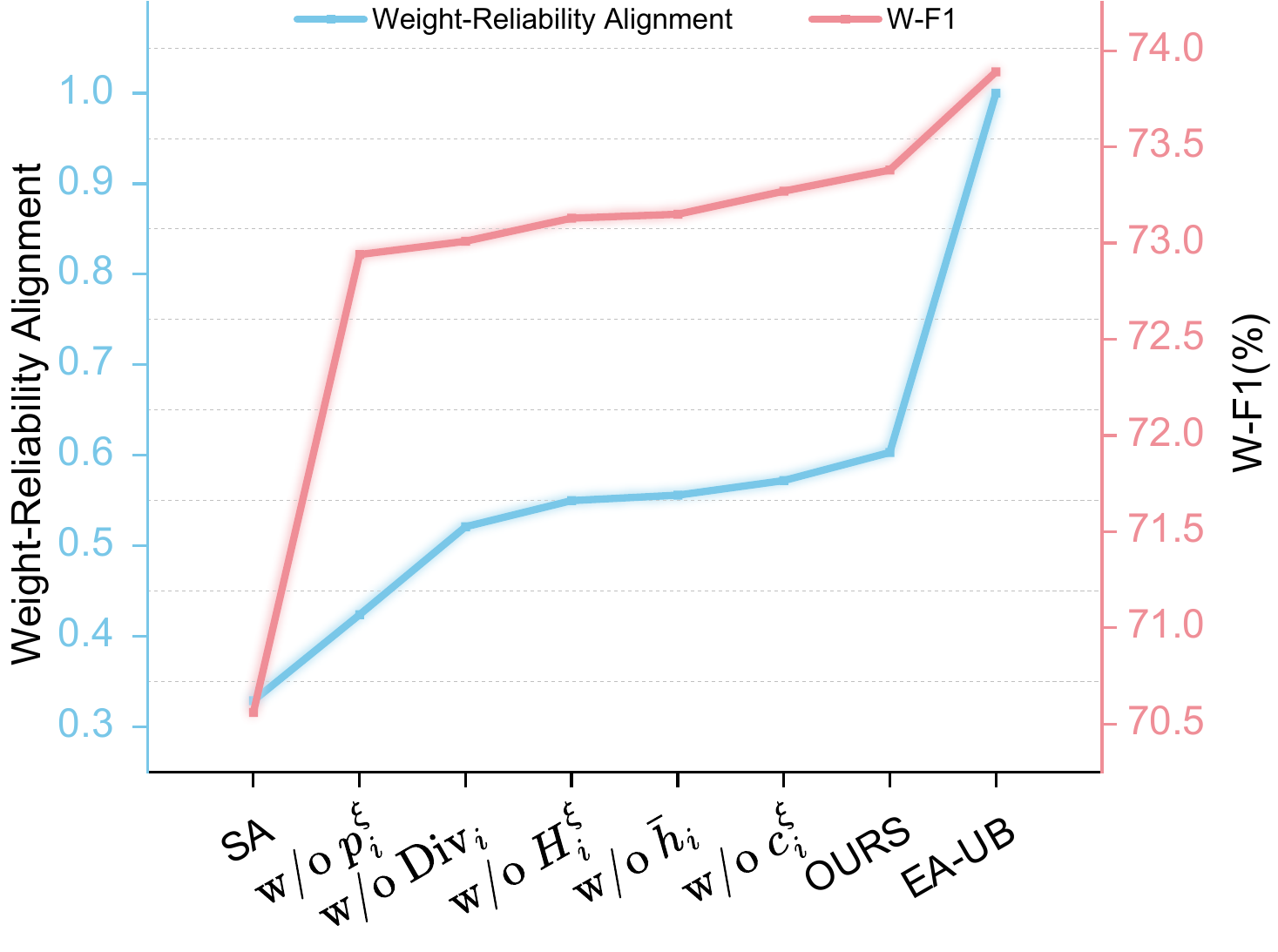}
        \caption{IEMOCAP} 
        \label{fig:sub_iemocap}
    \end{subfigure}
    \hfill
    \begin{subfigure}[b]{0.49\linewidth}
        \centering
        \includegraphics[width=\linewidth]{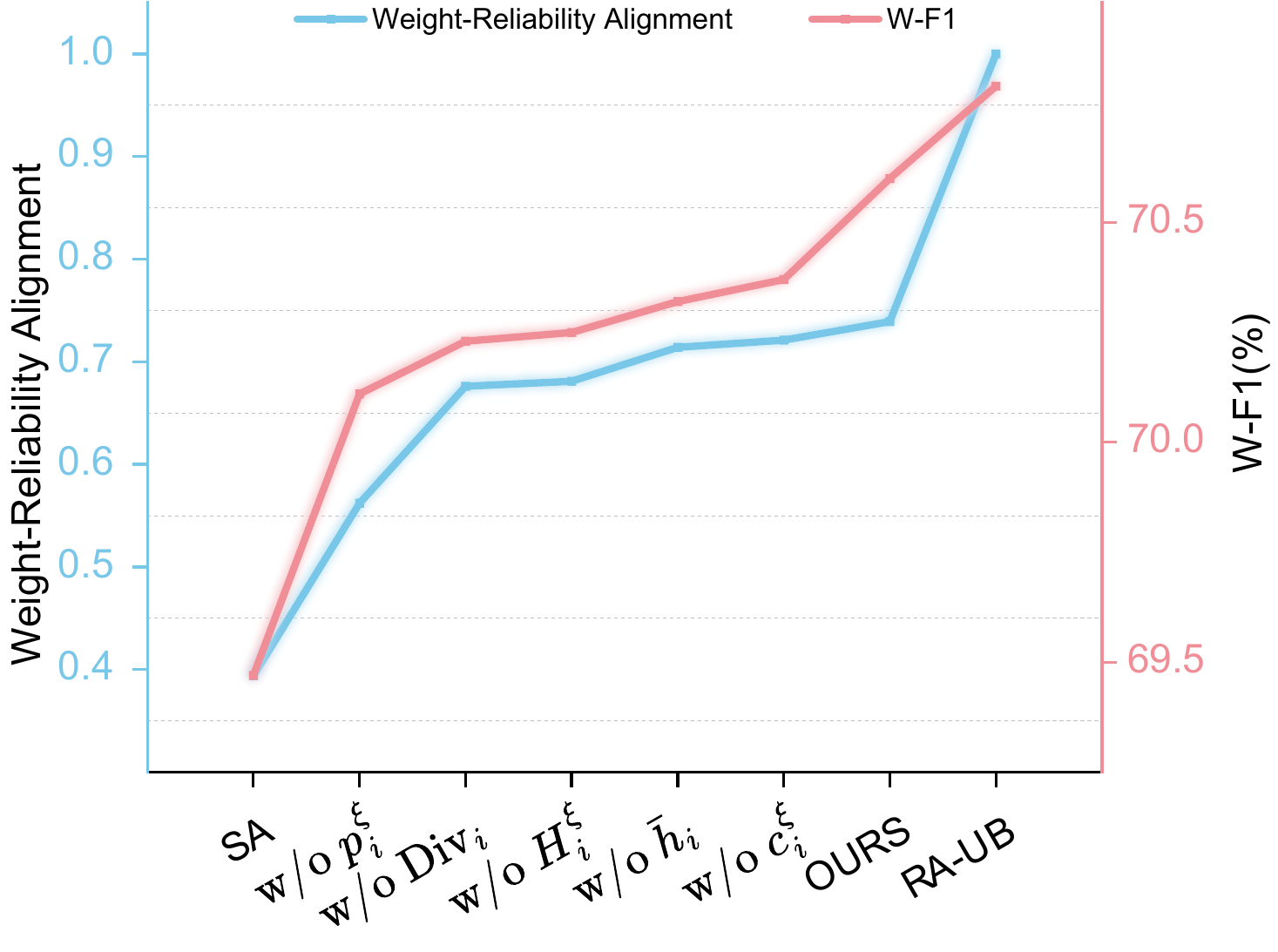}
        \caption{MELD} 
        \label{fig:sub_meld}
    \end{subfigure}
    
\caption{Weight–Reliability Alignment and Performance.}
    \label{fig:divergence}
\end{figure}
\textbf{Analysis of Weight-Reliability Alignment.} Figure~5 shows that modality weights increase with predicted relative reliability. We define Weight-Reliability Alignment (WRA) as the proportion of samples whose learned weight ranking matches the true reliability ranking given by $p_i^\xi(y_i)$. To assess the benefit of such alignment, we construct a Reliability-Aligned Upper Bound (RA-UB) by forcing the modality-weight ranking to follow this reference. As shown in Figure~6, RA-UB achieves the highest W-F1, while MiD-ERC outperforms the self-attention baseline in both WRA and W-F1 on the two datasets. Removing any input to the reliability estimator reduces both metrics, and the input ablations consistently show that higher WRA corresponds to higher W-F1, further supporting the effectiveness of divergence-modulated reliability weighting.\par
\textbf{Analysis of threshold $\theta$.} Figure 7 presents the sensitivity analysis of $\theta$. Both datasets achieve the best performance with a threshold of 0.05. We therefore adopt 0.05 as the default threshold.
\begin{figure}[!t]
    \centering
    \includegraphics[width=\linewidth]{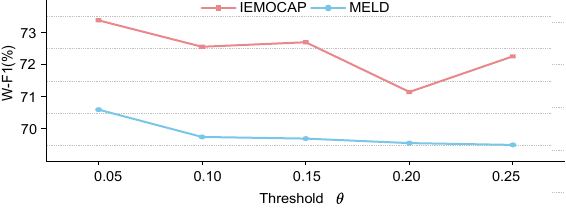}
    \caption{Sensitivity to the divergence threshold $\theta$.}
    \label{fig:orson_drawio}
\end{figure}
\subsection{Case Study}
\begin{figure}[!t]
    \centering
    \includegraphics[width=\linewidth]{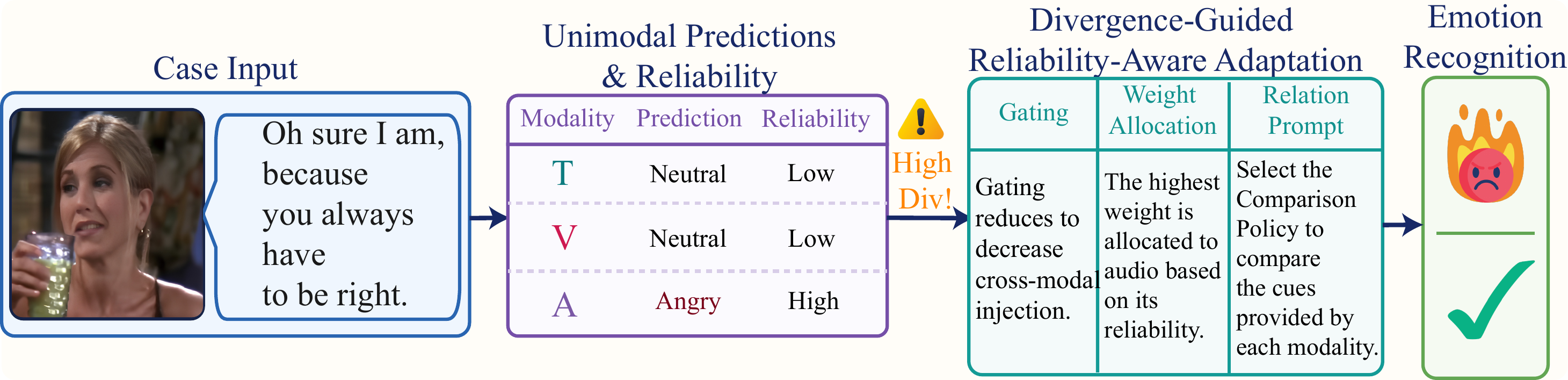}
    \caption{ Case study of MiD-ERC on a conflict sample.}
    \label{fig:orson_drawio}
\end{figure}
Figure 8 shows a high-divergence case where text and visual cues predict Neutral with low reliability, while audio predicts Angry with high reliability. MiD-ERC reduces cross-modal injection, prioritizes audio, and selects the Comparison Policy, leading to the correct recognition of Angry.
\section{Conclusion}
In this paper, we propose MiD-ERC, a divergence-centric MERC framework that models heterogeneous cross-modal inconsistencies as predictive disagreement among unimodal emotion distributions. It combines normalized inter-modal predictive divergence with sample-specific modality reliability to coordinate divergence-guided cross-modal gating, divergence-modulated reliability weighting, and divergence-conditioned adaptive LLM prompting. Trained end-to-end using only standard emotion annotations, MiD-ERC requires no conflict labels, auxiliary data construction, or external retrieval modules. Experiments on IEMOCAP and MELD show state-of-the-art performance, nearly half the inference cost, and larger gains on high-divergence samples.
\bibliography{aaai2027}
\appendix
\section{Detailed Method Formulation}
\subsection{Speaker-aware Causal GCN}
For each modality $\xi \in \{t,a,v\}$, we construct a modality-specific
utterance graph, where each utterance corresponds to one node. Given a
dialogue with $T$ utterances, the node set of modality $\xi$ is defined as
\begin{equation}
\mathcal{V}^{\xi}=\{v_1^\xi,v_2^\xi,\ldots,v_T^\xi\}
\end{equation}
The initial representation of node $i$ is the output of the preceding
temporal encoder:
\begin{equation}
r_i^\xi \in \mathbb{R}^{d}
\end{equation}

Different from DialogueGCN, which allows information propagation from both
past and future utterances, we construct a causal window graph. For target
node $i$, only previous utterances within a fixed historical window are
considered:
\begin{equation}
\mathcal{N}_i =
\{j \mid \max(1,i-W) \leq j < i\}
\end{equation}
where $W$ is the past window size. This causal construction prevents the
representation of the current utterance from accessing future information,
making it compatible with autoregressive LLM reasoning.

To distinguish intra-speaker and inter-speaker dependencies, we define
\begin{equation}
\delta_{ji} =
\begin{cases}
1, & s_j=s_i,\\
0, & s_j\neq s_i,
\end{cases}
\end{equation}
where $s_j$ and $s_i$ denote the speakers of utterances $j$ and $i$,
respectively. We also encode the relative temporal distance as:
\begin{equation}
d_{ji}=\frac{i-j}{W}
\end{equation}

For each directed edge $j \rightarrow i$, we concatenate the source
representation, target representation, speaker relation, and relative
distance:
\begin{equation}
\mathbf{e}_{ji}^{\xi}
=
\left[
r_j^\xi;
r_i^\xi;
\delta_{ji};
d_{ji}
\right]
\end{equation}
A two-layer MLP is then used to compute the unnormalized attention score:
\begin{equation}
a_{ji}^\xi = \mathrm{MLP}_{edge}^{\xi}
\left(\mathbf{e}_{ji}^{\xi}\right)
\end{equation}
The normalized attention weight is given by
\begin{equation}
\alpha_{ji}
=
\frac{
\exp(a_{ji}^\xi)
}{
\sum_{k\in \mathcal{N}_i}
\exp(a_{ki}^\xi)
}
\end{equation}

For message transformation, we use separate linear mappings for intra-speaker
and inter-speaker edges:
\begin{equation}
m_{ji}^\xi
=
\delta_{ji}\,
\mathrm{Linear}_{intra}^{\xi}(r_j^\xi)
+
(1-\delta_{ji})\,
\mathrm{Linear}_{inter}^{\xi}(r_j^\xi)
\end{equation}
The intra-speaker transformation captures emotional continuity of the same
speaker, while the inter-speaker transformation models emotional influence
across different speakers.

For the target node $i$, the self-projection term is computed as
\begin{equation}
u_i^\xi = \mathrm{Linear}_{self}^{\xi}(r_i^\xi)
\end{equation}
The speaker-aware historical context is then aggregated by
\begin{equation}
\bar{m}_i^\xi
=
\sum_{j\in \mathcal{N}_i}
\alpha_{ji} m_{ji}^\xi
\end{equation}
Finally, the updated node representation is obtained by
\begin{equation}
h_i^\xi
=
\mathrm{LayerNorm}
\left(
r_i^\xi
+
\mathrm{Dropout}
\left(
u_i^\xi + \bar{m}_i^\xi
\right)
\right)
\end{equation}
The resulting $h_i^\xi$ serves as the speaker-aware causal representation of
utterance $i$ under modality $\xi$.
\subsection{Connection to Bayesian Causal Inference}
Our divergence-modulated fusion is motivated by recent Bayesian causal inference work on communicative multisensory integration. This line of work suggests that multimodal integration is not a forced process; instead, observers infer whether different sensory signals should be attributed to a common communicative cause or treated as separate sources of evidence. In particular, prior expectations about whether multimodal signals are structurally and pragmatically coherent can arbitrate between integration and segregation.

Following this principle, we regard the inter-modal divergence $\mathrm{Div}_i$ as a computational proxy for the uncertainty of whether textual, acoustic, and visual cues reflect the same latent emotional cause. When $\mathrm{Div}_i$ is small, the unimodal prediction distributions are consistent, suggesting that the three modalities are likely to support the same underlying emotional state. In this case, MiD-ERC uses a lower fusion temperature, producing a sharper reliability-weighted distribution so that the most reliable modality can be emphasized more decisively.

In contrast, when $\mathrm{Div}_i$ is large, the modalities provide conflicting emotional evidence. Such conflicts may arise from sarcasm, concealed emotion, modality-specific noise, or contextual ambiguity. Under this condition, assigning an overly dominant weight to a single modality may prematurely suppress other informative cues. Therefore, MiD-ERC increases the fusion temperature to smooth the modality weights, preserving multiple competing pieces of evidence instead of collapsing them into a single modality-dominated representation.
\section{Prompt Design}
\subsection{Static Prompt Template}
To adapt multimodal ERC to the LLM input format, we use the following static
prompt template for each target utterance. The placeholders
\texttt{<MultiHere>}, \texttt{<TextHere>}, \texttt{<AudioHere>}, and
\texttt{<FrameHere>} denote the projected fused multimodal, textual, acoustic,
and visual embeddings, respectively. The placeholder
\texttt{<RelationHere>} is dynamically filled with the selected modality
relation policy according to the divergence level. Details are shown in Figure 9. The static prompt defines the fixed input structure, while
\texttt{<RelationHere>} is dynamically filled according to the inter-modal
divergence level. For low-divergence samples, it is filled with the
consistency policy; for high-divergence samples, it is filled with the
conflict policy.
\begin{figure}[!t]
\centering
\begin{promptbox}{Static prompt template for MiD-ERC.}
###Human:
The fused multimodal feature is:
<Multi><MultiHere></Multi>.
The modality relation policy is:
<Relation><RelationHere></Relation>.
The textual feature is:
<Text><TextHere></Text>.
The audio feature is:
<Audio><AudioHere></Audio>.
The visual feature is:
<Video><FrameHere></Video>.
The dialogue context and task are:
<Dialogue>
Now you are expert of sentiment and emotional analysis.
The following conversation noted between '### ###'
involves several speakers.
###
{Speaker_1}: "{utterance_1}"
{Speaker_2}: "{utterance_2}"
...
{Target_Speaker}: "{target_utterance}"
###
Please select the emotional label of
<{Target_Speaker}: "{target_utterance}">
from <{label_options}>:
</Dialogue>
###Assistant:
\end{promptbox}
\caption{Static prompt template used to formulate each target utterance as an LLM input.}
\label{fig:static_prompt}
\end{figure}
\subsection{Relation Prompt Template}
The placeholder \texttt{<RelationHere>} in the static prompt is dynamically
replaced by one of the relation policies shown in Figure 10 according to the
divergence threshold $\theta$. Specifically, the consistency policy is used
when $\mathrm{Div_i} < \theta$, while the conflict policy is used when
$\mathrm{Div_i} \geq \theta$.
\begin{figure}[!t]
\centering
\begin{prompttextbox}{Relation prompt template for MiD-ERC.}

\noindent\textbf{[Consistency relation policy]}\\
The three modalities of this sample are consistent. Use the dialogue context
and \ptag{Multi} as the primary evidence for predicting the target speaker's
emotion. \ptag{Text}, \ptag{Audio}, and \ptag{Video} provide supporting
modality-specific evidence. Because the modalities are consistent, trust
\ptag{Multi} as the main multimodal representation.

\vspace{0.8em}

\noindent\textbf{[Conflict relation policy]}\\
The three modalities of this sample are conflicting. Do not rely only on
\ptag{Multi} or on the most confident single modality. Compare \ptag{Text},
\ptag{Audio}, and \ptag{Video} with the dialogue context to resolve the
conflict. The conflict may come from sarcasm, concealed emotion, or modality
noise. Predict the final emotion from the most coherent explanation of the
dialogue context and modality-specific evidence.

\end{prompttextbox}
\caption{Relation prompt templates used for dual-prompt conflict-aware LLM reason-
ing mechanism.}
\label{fig:relation_prompt}
\end{figure}
\section{Dataset and Input Construction}
\subsection{Dataset Details}
IEMOCAP is a widely used benchmark dataset for multimodal emotion recognition developed by the University of Southern California (USC). It consists of dyadic conversational interactions performed by professional actors in a controlled laboratory setting, integrating audio, video, text transcripts, and high-precision motion capture data. The dataset provides utterance-level annotations for both discrete emotion categories and continuous emotional dimensions. Due to its high-quality multimodal alignment and rigorous experimental design, it has become a cornerstone for fine-grained multimodal emotion computation research.\par
MELD is a large-scale, multimodal emotion dataset derived from the classic TV sitcom Friends, specifically designed for multi-party conversational scenarios. It encompasses synchronized audio, video, and textual modalities, with utterance-level annotations for both discrete emotions and coarse-grained sentiment orientations. Compared to controlled laboratory environments, MELD features naturalistic spoken language, rich social contexts, and frequent speaker turns in multi-party interactions, making it one of the most representative benchmarks in MERC.
\subsection{Construction of Multimodal Instruction Datasets}
We construct the multimodal instruction datasets for both IEMOCAP and MELD
using the same preprocessing pipeline. For each dataset, we align the original
dialogue annotations with pre-extracted multimodal features. The original
annotation files provide utterance-level speaker identities, textual contents,
emotion labels, and the official train, validation, and test dialogue split, while
the multimodal feature files provide utterance-level textual, acoustic, and
visual features.\par
For each target utterance, we build an instruction-style prompt using a fixed
historical window. The context is formatted with speaker identifiers and the
model is asked to select the emotion label of the target utterance from the
candidate label set of the corresponding dataset. The instruction template is
shown in Figure 11.
\begin{figure}[t]
\centering
\begin{promptbox}{Instruction template for dataset construction.}
Now you are expert of sentiment and emotional analysis.
The following conversation noted between `\#\#\# \#\#\#' involves several speakers.
\#\#\#
\{Speaker\_1\}: ``\{utterance\_1\}''
\{Speaker\_2\}: ``\{utterance\_2\}''
...
\{Target\_Speaker\}: ``\{target\_utterance\}''
\#\#\#
Please select the emotional label of
<\{Target\_Speaker\}: ``\{target\_utterance\}''>
from <\{label\_options\}>:
\end{promptbox}
\caption{Instruction template used for constructing IEMOCAP and MELD samples.}
\label{fig:dataset_instruction_template}
\end{figure}\par
For multimodal feature storage, we use the pre-extracted multimodal features
from CFN-ESA. The utterance-level textual, acoustic, and visual features of
each dialogue are saved as a dialogue-level \texttt{.npz} bundle, and each
JSON sample records the corresponding bundle path and utterance index for
feature retrieval. For IEMOCAP, each bundle contains four textual feature
streams, \texttt{textf1}, \texttt{textf2}, \texttt{textf3}, and
\texttt{textf4}, together with \texttt{audio} and \texttt{visual}; their
feature dimensions are $1024$, $1582$, and $342$, respectively. MELD follows
the same construction pipeline with its dataset-specific CFN-ESA features.\par
Each utterance is converted into one JSON sample. The main fields include the
sample ID, dataset name, conversation ID, turn ID, utterance ID, speaker ID,
raw text, instruction input, target emotion label, label index, and
multimodal feature metadata. The multimodal metadata records the path to the
dialogue-level feature bundle, the utterance-level feature index, and the
keys used to retrieve textual, acoustic, and visual features. An example
schema is shown in Figure 12:
\begin{figure}[t]
\centering
\begin{promptbox}{JSON schema of a processed sample.}
{
  "sample_id": "{conv_id}_{turn_id}",
  "dataset": "{dataset_name}",
  "conv_id": "{conv_id}",
  "turn_id": 0,
  "utterance_id": "{utterance_id}",
  "speaker": "{speaker}",
  "text": "{target_utterance}",
  "input": "{instruction_prompt}",
  "target": "{emotion_label}",
  "label_id": 0,
  "multimodal": {
    "features_path": "{conv_id}.npz",
    "features_index": 0,
    "text_keys": ["textf1", "textf2", "textf3", "textf4"],
    "audio_key": "audio",
    "visual_key": "visual"
  }
}
\end{promptbox}
\caption{JSON schema of each processed multimodal instruction sample.}
\label{fig:json_schema}
\end{figure}
\section{Implementation Details}
We use Qwen2.5-7B-Instruct as the base LLM. During training, the parameters of the base LLM are frozen, and only LoRA adapters and the trainable multimodal modules are updated. The trainable multimodal modules include the multimodal projection layers, unimodal context encoders, speaker-aware causal GCN, reliability heads, divergence gate, and fusion layers. We use bfloat16 precision for LLM computation. For parameter-efficient adaptation, we apply LoRA to the frozen LLM. The LoRA rank, scaling factor, and LoRA dropout are set to $16$, $32$, and $0.05$, respectively. It should be noted that this dropout only refers to the LoRA adapters. The dropout used in the trainable non-LLM modules is set separately: $0.2$ for IEMOCAP and $0.3$ for MELD. The divergence-gated cross-modal attention module uses $4$ attention heads. The historical context window is the same as Casual-ERC, which is set to $8$ for IEMOCAP and $6$ for MELD, the graph window W is set to 8 for IEMOCAP and 6 for MELD too.\par 
For divergence-modulated fusion, we set the relation divergence threshold to $\theta=0.05$, the base temperature to $\tau_0=0.5$, and the divergence temperature coefficient to $\alpha=2.0$. The loss weights are set to $\lambda_{uni}=0.8$, $\lambda_{rel}=0.15$, and $\lambda_{ent}=0.05$. The main classification loss has a weight of $1.0$. These values are all derived from extensive sensitivity experiments. We use the AdamW optimizer combined with a linear warmup cosine scheduler to train the model. The initial learning rate is $2\times10^{-5}$, the minimum learning rate is $1\times10^{-6}$, the warmup learning rate is $1\times10^{-7}$, and the number of warmup steps is $1000$. The weight decay is set to $0.05$. All experiments are trained for $10$ epochs with $1000$ iterations per epoch on a single RTX 5880 ADA. The training batch sizes for IEMOCAP and MELD are $2$ and $3$, respectively.
\section{Additional Experiments}
\subsection{Different LLM Backbones}
Table 5 shows that MiD-ERC maintains consistent performance across different LLM backbones. Llama-3.1-8B-Instruct achieves the best results, while the overall variations remain within 0.71\% on IEMOCAP and 0.21\% on MELD, demonstrating that the effectiveness of MiD-ERC is not tied to a specific backbone. In the main experiments, although Qwen2.5-7B-Instruct does not yield the highest performance, we adopt it as the default backbone to maintain consistency with the backbone settings used by most LLM-based baselines, thereby ensuring a fair and controlled comparison.
\begin{table}[!t]
  \centering
  \label{tab:ablation}
  \resizebox{\columnwidth}{!}{%
  \begin{tabular}{lcccc} 
    \toprule
    \multirow{2}{*}[-1ex]{Methods} & \multicolumn{2}{c}{IEMOCAP} & \multicolumn{2}{c}{MELD} \\
    \cmidrule(lr){2-3} \cmidrule(lr){4-5}
    & Acc & W-F1 & Acc & W-F1 \\
    \midrule
    Qwen2.5-7B-Instruct   & 73.44 & 73.38 & 71.79 & 70.60 \\
   Llama-3.1-8B-Instruct                 & 73.72          & 73.64         & 71.95         & 70.81         \\
     Qwen3-8B           & 73.01        & 72.97       & 71.83     & 70.76         \\
    \bottomrule
  \end{tabular}%
  }
  \caption{Ablation study on IEMOCAP and MELD.}
\end{table}
\subsection{Parameter-Shared Context Encoders}
Table~\ref{tab:shared_encoder} examines the effect of sharing the DialogueRNN and speaker-aware GCN parameters across modalities. Although parameter sharing reduces the model size by 1.87\% on IEMOCAP and 0.71\% on MELD, it causes W-F1 drops of 4.20\% and 3.09\%, respectively. This indicates that the limited parameter saving does not compensate for the loss in performance. Since textual, acoustic, and visual signals exhibit different temporal and speaker-dependent patterns, modality-specific context encoders are necessary to preserve heterogeneous emotional cues and provide reliable unimodal evidence for subsequent divergence estimation and reliability-aware fusion.

\begin{table}[t]
\centering
\caption{Effect of parameter sharing in unimodal context encoders. Relative changes are calculated with respect to MiD-ERC.}
\label{tab:shared_encoder}
\setlength{\tabcolsep}{3pt}
\renewcommand{\arraystretch}{1.05}
\resizebox{\columnwidth}{!}{
\begin{tabular}{llccc}
\toprule
Dataset & Variant & Params. & Acc & W-F1 \\
\midrule
\multirow{3}{*}{IEMOCAP}
& MiD-ERC
& 147.826M
& 73.44
& 73.38 \\
& Shared Context Encoders
& 145.056M
& 70.61
& 70.30 \\
& \textit{Relative Change}
& $\downarrow 1.87\%$
& $\downarrow 3.85\%$
& $\downarrow 4.20\%$ \\
\midrule
\multirow{3}{*}{MELD}
& MiD-ERC
& 136.733M
& 71.79
& 70.60 \\
& Shared Context Encoders
& 135.768M
& 70.11
& 68.42 \\
& \textit{Relative Change}
& $\downarrow 0.71\%$
& $\downarrow 2.33\%$
& $\downarrow 3.09\%$ \\
\bottomrule
\end{tabular}
}
\end{table}
\end{document}